\documentclass{article}

\usepackage[normalem]{ulem}

\newcommand{\revised}[1]{%
\ifx\highlightrevisions\undefined{#1}%
\else\textcolor{red}{#1}%
\fi}

\newcommand{\rnum}[1]{%
\ifx\showreviewercommentnum\undefined%
\else{[\bf{#1}] }%
\fi}

\newcommand{\erased}[1]{%
\ifx\showerased\undefined%
\else{\sout{#1}}%
\fi}

\usepackage{authblk}
\usepackage{relsize}
\usepackage{gensymb}
\usepackage{url}
\usepackage{bm}
\usepackage{amsmath}
\usepackage{cite}
\usepackage{graphicx}
\usepackage{multirow}
\usepackage{booktabs}
\usepackage[font=small]{caption}
\usepackage{mathtools}
\usepackage{algorithm}
\usepackage{algorithmic}
\usepackage{amsthm,amsfonts,amsmath,amssymb}
\usepackage{graphicx,url}
\usepackage{helvet}
\usepackage{tabularx}
\usepackage{color}
\usepackage{float}
\usepackage[colorlinks=true,linkcolor=blue,citecolor=magenta]{hyperref}
\usepackage{array}

\usepackage{enumerate}
\usepackage[shortlabels]{enumitem}

\theoremstyle{definition}

\newcommand{ \prox} {\mathcal{P}}
\bmdefine\magn{m}
\bmdefine\phase{p}
\bmdefine\wrap{w}
\bmdefine\y{y}
\bmdefine\x{x}
\bmdefine\resid{r}

\newcommand{ \wraps } { \mathcal{W}}
\newcommand{ \minimize }[1] { \underset{#1}{\text{minimize }}}

\let\citeleft=(
\let\citeright=)

\title{\vspace{-2cm}General Phase Regularized Reconstruction using Phase Cycling}
\author{Frank Ong$^1$, Joseph Cheng$^2$, and Michael Lustig$^1$}

\affil{
$^1$Department of Electrical Engineering and Computer Sciences, \\
    University of California, Berkeley, California. \\
    $^2$Department of Radiology, \\
    Stanford University, Stanford, California.}

\begin{document}

\maketitle

\vfill

\noindent
\textit{Running head:} General Phase Regularized Reconstruction using Phase Cycling

\vspace{0.5cm}
\noindent
\textit{Address correspondence to:} \\
  Michael Lustig \\
  506 Cory Hall \\
  University of California, Berkeley \\
  Berkeley, CA 94720 \\
  mlustig@eecs.berkeley.edu

\vspace{0.5cm}
\noindent
This work was supported by NIH R01EB019241, NIH R01EB009690, GE Healthcare, and NSF Graduate Fellowship.

\vspace{0.3cm}
\noindent
Approximate word count: 224 (Abstract) 3925 (body)\\

\vspace{0.3cm}
\noindent
Submitted to \textit{Magnetic Resonance in Medicine} as a Full Paper.\\

\vspace{0.3cm}
\noindent
Part of this work has been presented at the ISMRM Annual Conference 2014 and 2017.

\newpage
\section*{Abstract}

\noindent
\textbf{Purpose:} To develop a general phase regularized image reconstruction method, with applications to partial Fourier imaging, water-fat imaging and flow imaging.
\\

\noindent
\textbf{Theory and Methods:} The problem of enforcing phase constraints in reconstruction was studied under a regularized inverse problem framework. A general phase regularized reconstruction algorithm was proposed to enable \revised{\erased{the}various} joint reconstruction of partial Fourier imaging, water-fat imaging and flow imaging, along with parallel imaging (PI) and compressed sensing (CS). Since phase regularized reconstruction is inherently non-convex and sensitive to phase wraps in the initial solution, a reconstruction technique, named phase cycling, was proposed to \revised{\rnum{R2.12.1}\erased{enable}render} the overall algorithm invariant to phase wraps. The proposed method was applied to retrospectively under-sampled in vivo datasets and compared with state of the art reconstruction methods.
\\

\noindent
\textbf{Results:} Phase cycling reconstructions showed reduction of artifacts compared to reconstructions without phase cycling and achieved similar performances as state of the art results in partial Fourier, water-fat and divergence-free regularized flow reconstruction. Joint reconstruction of partial Fourier + water-fat imaging + PI + CS, and partial Fourier + divergence-free regularized flow imaging + PI + CS were demonstrated.
\\

\noindent
\textbf{Conclusion:} The proposed phase cycling reconstruction provides an alternative way to perform phase regularized reconstruction, without the need \revised{\rnum{R2.12.2}\erased{of performing}to perform} phase unwrapping. It is robust to the choice of initial solutions and \revised{\erased{enables}encourages} the joint reconstruction of phase imaging applications.


\newpage
\section{Introduction}

Phase variations in \revised{\rnum{R2.12.3}\erased{Magnetic Resonance Imaging (MRI)}MRI} can be attributed to a number of factors, including field inhomogeneity, chemical shift, fluid flow, and magnetic susceptibility. These phase structures are often the bases of many useful MRI applications: Image phase from $B_0$ inhomogenity is used for calibration and shimming. Others provide important physiological information in clinical imaging methods, such as chemical shift imaging, phase contrast imaging, and susceptibility imaging.

Phase structures in MRI provide opportunities in reconstruction to resolve undersampling artifacts or to extract additional information. For example, in partial Fourier imaging, smoothness of the phase images is exploited to reduce acquisition time by factors close to two. In water-fat imaging, \revised{\rnum{R2.12.4}chemical shift induced phase shifts are used to separate water and fat images, while smoothness in $B_0$ field inhomogeneity is exploited to prevent water-fat swaps.\erased{structures in $B_0$ field inhomogeneity are exploited to separate water and fat images.}} These methods have \revised{\rnum{R2.12.4}\erased{resulted in reduction of acquisition time while providing additional diagnostic information.}either reduced acquisition time as in the case of partial Fourier imaging, or provided more accurate diagnostic information as in the case of water-fat imaging.}

In this paper, we study the problem of exploiting these phase structures in a regularized inverse problem formulation. An inverse problem formulation allows us to easily incorporate parallel imaging~\cite{Pruessmann:1999cw, Griswold:2002gf} and compressed sensing~\cite{Lustig:2007cu} and utilize various phase regularizations for different applications. Our first contribution is to present a unified reconstruction framework for phase regularized reconstruction problems. \revised{\rnum{R2.12.5}\erased{and enable joint reconstruction of partial Fourier, water-fat imaging \revised{and / or} flow imaging applications.}In particular, using the same framework, we demonstrate the joint reconstruction of partial Fourier and water-fat imaging, along with parallel imaging (PI) and compressed sensing (CS), and the joint reconstruction of partial Fourier and divergence-free regularized flow imaging, along with PI + CS.}

Since phase regularized image reconstructions are inherently non-convex and are sensitive to phase wraps in the initial solution, we propose a reconstruction technique, named phase cycling, that enables phase regularized reconstruction \revised{\rnum{R.2.12.6}to be }robust to phase wraps. The proposed phase cycling technique is inspired by cycle spinning in wavelet denoising~\cite{Coifman:1995ji} and phase cycling in balanced steady state free precession. Instead of \revised{\rnum{R.2.12.7}\erased{unwraping}unwrapping} phase before or during the reconstruction, the proposed phase cycling makes the reconstruction invariant to phase wraps by cycling different initial phase \revised{\rnum{R2.12.8}\erased{solution}solutions} during the reconstruction. Our key difference with existing works is that the proposed phase cycling does not solve for the absolute phase in reconstruction, but rather averages out the effects of phase wraps in reconstruction. We provide experimental results showing its robustness to phase wraps in initial solutions.

\subsection*{Related Works}
\label{ssec:related}

Reconstruction methods utilizing phase structures were proposed for water-fat reconstruction~\cite{Reeder:2005be, Reeder:2003ib, Yu:2005kz, Doneva:2010it, Sharma:2011ca, Lu:2008ke, Hernando:2008jv, Reeder:2007iv, Eggers:2014cy}, $B_0$ field map estimation~\cite{Funai:2008gt}, partial Fourier reconstruction~\cite{Noll:2004dc, Haacke:1991hc, Bydder:2005bya, Li:2015fh} and divergence-free regularized 4D flow reconstruction~\cite{Loecher:2012uq, Ong:2013te, Santelli:2015go}. In addition, Reeder et al.~\cite{Reeder:2005bp} demonstrated the feasibility of jointly performing homodyne reconstruction and water-fat decomposition. Johnson et al.~\cite{Johnson:2010ff} presented results in jointly reconstructing water-fat and flow images.

A closely related line of work is the separate magnitude and phase reconstruction method~\cite{Fessler:2004hg, Zibetti:2010gx, Zhao:2012ed, Zibetti:2016cp}. In particular, Fessler and Noll~\cite{Fessler:2004hg} proposed using alternating minimization for separate magnitude and phase reconstruction\revised{, but their method remained sensitive to phase wraps in initial solution.} Zibetti et al.~\cite{Zibetti:2010gx} and Zhao et al.~\cite{Zhao:2012ed} achieved robustness to phase wraps in initial solution by designing a regularization function that resembles the finite difference penalty, and is periodic in the phase. \revised{The regularization function proposed by Zhao et al. differs from the one proposed by Zibetti et al. in that it is edge-preserving using the Huber loss function.} One limitation of these methods is that they do not support general magnitude and phase operators, or arbitrary phase regularization. In particular, they cannot be applied to water-fat image reconstruction and flow image reconstruction with general velocity encoding, or regularization functions other than finite difference. \revised{This restriction to finite difference penalty can lead to well-known staircase artifacts as shown in Figure~\ref{fig:brain_compare}.} In contrast, our proposed method can be applied to general phase imaging methods and supports arbitrary phase regularization as long as its proximal operator can be computed. \revised{\rnum{R.2.12.12}This} allows us to enforce application dependent regularization, such as wavelet sparsity penalty and divergence-free constraint for flow imaging.


\section{Theory}

\subsection*{Forward Model and Applications}
\label{sec:problem} 

In this section, we describe the forward models for partial Fourier imaging, water-fat imaging and flow imaging. We then show that these phase imaging methods can be described using a general forward model and hence can be \revised{\rnum{R2.12.11}\erased{jointly reconstructed} combined} within the same framework. \revised{\rnum{R.12.11}For example, in the experimental sections, we combine partial Fourier with water-fat imaging, and partial Fourier with divergence-free regularized flow imaging, under the same framework.} Figure~\ref{fig:model} provides illustrations of the forward models of these applications.

\begin{figure}[!ht]
  \centering
  \includegraphics[width=\linewidth]{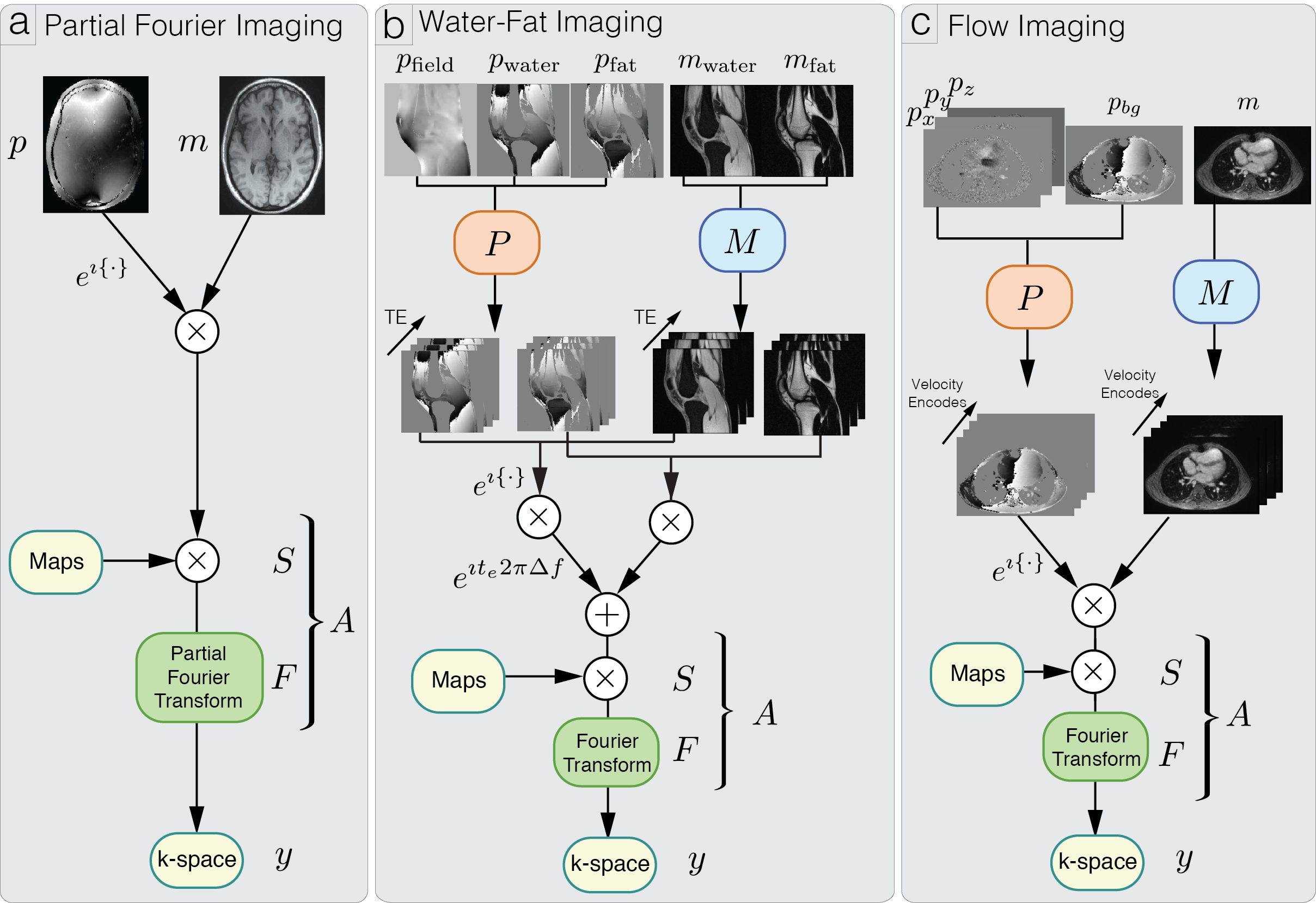}
  \caption{Illustration of forward models y $= A (M m \cdot e^{\imath  Pp})$ for partial Fourier imaging, water fat imaging and flow imaging.}. 
  \label{fig:model}
\end{figure}

\subsubsection*{Partial Fourier Imaging}

In partial Fourier imaging, a contiguous portion of k-space is not observed and the k-space data is observed through the partial Fourier operator $F$ and sensitivity operator $S$.
Let $m$ be the magnitude images and $p$ be the phase images, our forward model is given by:
\begin{equation}
	\begin{aligned}
          y = F S \left(m \cdot e^{\imath  p } \right) + \eta
	\end{aligned}
	\label{eq:pf_d}
\end{equation}
where $\cdot$ is the element-wise multiplication operator, \revised{$\imath$ denotes $\sqrt{-1}$, $e^{\imath p}$ denotes the element-wise exponential operator of the vector $\imath p$}, \revised{\rnum{R3.6.1}\erased{$z$}$\eta$} is a complex, white Gaussian noise vector, and $y$ represents the acquired k-space data.

Traditional partial k-space reconstruction methods assume smoothness of the phase. Therefore, in our formulation, we directly impose smoothness constraint on the phase image $p$.\revised{\rnum{R3.6.2}\erased{to enforce conjugate symmetry in k-space.}} Sparsity of the magnitude image can also be exploited for compressed sensing applications.

\subsubsection*{Water-fat Imaging}

In water-fat imaging, we are interested in reconstructing water and fat images with \revised{\rnum{R2.12.13}\erased{different off-resonance frequencies}different chemical shift induced off-resonance, but the same $B_0$ inhomogeneity induced off-resonance.} In order to resolve these images, k-space measurements are acquired over multiple echo times. 

Concretely, let $E$ be the number of echo times, $m_\text{water}$ and $m_\text{fat}$ be the water and fat magnitude images respectively, $p_\text{water}$ and $p_\text{fat}$ be the water and fat phase images respectively, $\Delta f$ be the frequency difference between water and fat under the single-peak model, and $p_\text{field}$ be the $B_0$ field inhomogenity. We denote $F_e$ as the Fourier sampling operator for each echo time $t_e$. Our forward model is given by:
\begin{equation}
	\begin{aligned}
		 y_e &= F_e  S \left( \left(m_\text{water} e^{\imath  p_\text{water}} + m_\text{fat} e^{\imath  p_\text{fat}} e^{\imath  t_e 2 \pi \Delta f} \right) e^{\imath  t_e p_\text{field}} \right) + \eta_e &\text{for } e = 1, \hdots, E
	\end{aligned}
	\label{eq:cs_d}
\end{equation}
We note that the water and fat images have independent phase images $p_\text{water}$ and $p_\text{fat}$, which are often implicitly captured in existing water-fat separation methods by representing the magnitude images as complex images. These phase images can be attributed to RF pulse with spectral variation and $B_1$ field inhomogeneity. We explicitly represent these phase images because they can be regularized as spatially smooth when partial Fourier imaging is incorporated~\cite{Reeder:2005bp}.

In order to separate the components, the first-order field inhomogeneity \revised{\rnum{R2.12.14}\erased{$p$}$p_\text{field}$} is regularized to be spatially smooth. In addition, sparsity constraints can be imposed on magnitude images for compressed sensing applications.

\subsubsection*{Flow Imaging}

In three-dimensional phase contrast imaging, we are interested in reconstructing phase images with three-dimensional velocity information. Concretely, we define $p = (p_{bg}, p_x, p_y, p_z)^\top$ to be the \revised{\erased{reference}background} phase image and velocity images along three spatial dimensions $(x, y, z)$ respectively. \revised{\rnum{R2.12.16}The background phase includes the $B_0$ field inhomogeneity, and chemical shift induced phase.} We also let $V$ be the number of velocity encodes, and $P_v$ be the velocity encoding vector for velocity encode \revised{\rnum{R2.12.15}\erased{$e$}$v$}. For example, the four point balanced velocity encoding~\cite{Pelc:2005jj} vectors, ignoring scaling, have the form: \revised{\rnum{R1.5}\erased{$P_e = (+1, \pm 1, \pm 1, \pm 1)$.}}
\begin{equation}
  \begin{aligned}
    P_1 &= (+1, -1, -1, -1)\\
    P_2 &= (+1, +1, +1, -1)\\
    P_3 &= (+1, +1, -1, +1)\\
    P_4 &= (+1, -1, +1, +1)\\
  \end{aligned}
\end{equation}

Then, our forward model is given by:
\begin{equation}
	\begin{aligned}
		y_v &= F_v S \left( m \cdot e^{\imath  P_v p} \right)   + \eta_v &\text{for } v = 1, \hdots, V
	\end{aligned}
	\label{eq:flow_d}
\end{equation}
Since blood flow is incompressible and hence divergence-free, the velocity images $p_x, p_y, p_z$ can be constrained to be divergence-free to provide more accurate flow rates~\cite{Loecher:2012uq, Ong:2013te, Santelli:2015go}. Smoothness constraint can also be imposed on the \revised{\erased{reference}background} phase image to improve flow accuracy.

\subsubsection*{General Forward Model}

With the right representations shown in Appendix~\ref{sec:water_fat_flow_general}, the above phase imaging applications can all be described using the following unified forward model:
\begin{equation}
	\begin{aligned}
		 y = A \left(M m \cdot e^{\imath  P p } \right) + \eta
	\end{aligned}
	\label{eq:model}
\end{equation}
where $y$ is the acquired k-space measurements, $m$ contains the magnitude images, $p$ contains the phase images, $\eta$ is a white Gaussian noise vector, $A$ is the forward operator for the complex images, $M$ is the forward operator \revised{\rnum{R3.6.3}for }the magnitude image, $P$ is the forward operator for the phase image and $\cdot$ is the point-wise multiplication operator. We note that both $m$ and $p$ are real-valued.

\subsection*{Objective function}
\label{ssec:obj}

To reconstruct the desired magnitude images $m$ and phase images $p$, we consider the following regularized least squares function:
\begin{equation}
	\begin{aligned}
		\underbrace{ \frac{1}{2} \left\| y - A \left(M m \cdot e^{\imath  P p } \right)  \right\|_2^2}_{\text{Data consistency } f(m, p)} + \underset{\text{Regularization }  g(m, p)}{ \underbrace{  \vphantom{\frac{1}{2}}g_m ( m ) + g_p (p)}}\\
	\end{aligned}
	\label{eq:obj}
\end{equation}
where $g_m$ and $g_p$ are regularization functions for magnitude and phase respectively. For notation, we split our objective function into a data consistency term $f(m,p)$ and a regularization term $g(m,p)$.

We note that our objective function is non-convex \revised{\rnum{R2.2}regardless of the application-dependent linear operators $A$, $M$, and $P$, because} the exponential term has periodic saddle points in the data consistency function with respect to $p$. In addition, the forward model is bi-linear in the magnitude and complex exponential variables.

\subsection*{Algorithm}
\label{ssec:algorithm} 

Since the phase regularized reconstruction problem is non-convex in general, \revised{\rnum{R2.2}finding the global minimum is difficult. In fact, finding a local minimum is also difficult with conventional gradient-based iterative methods as saddle points have zero gradients. Instead we aim for} a monotonically \revised{\rnum{R2.12.18}\erased{descent}descending} iterative algorithm \revised{\erased{is desired}} to ensure the reconstructed result reduces the objective function. In particular, we perform alternating minimization with respect to the magnitude and phase images separately and use the proximal gradient method~\cite{Combettes:2011iq, Parikh:2013vb} for each sub-problem, which is guaranteed to \revised{\rnum{R2.12.18}\erased{descent}descend} with respect to the objective function in each iteration. \revised{\rnum{R2.2}Since the objective function is reduced in each iteration as long as the gradient is non-zero, the algorithm eventually converges to a stationary point with zero gradient, which can be either a local minimum or a saddle point, assuming the initialization is not exactly at a local maximum.} A high level illustration is shown in Figure~\ref{fig:alg_overall}. 

Concretely, applying proximal gradient descent on our objective function~\eqref{eq:obj}, we perform a gradient descent step with respect to the data consistency function $f(m, p)$ and a proximal operation with respect to the regularization function $g(m, p)$. Then, we obtain the following update step for magnitude images $m$ with fixed phase images $p$ at iteration $n$:
\begin{equation}
	\begin{aligned}
		m_{n+1} &= \prox_{\alpha_n g_m} \left( m_n - \alpha_n \nabla_{m} f(m_n, p) \right)\\
	\end{aligned}
	\label{eq:mag_update}
      \end{equation}
and the following update step for phase images $p$ with fixed magnitude images $m$ at iteration $n$:
\begin{equation}
	\begin{aligned}
		p_{n+1} &= \prox_{\alpha_n g_p} \left( p_n - \alpha_n \nabla_{p} f(m, p_n) \right)\\
	\end{aligned}
	\label{eq:phase_update}
\end{equation}
where $\prox_{g}(x) = \text{argmin}_z \frac{1}{2} \| z - x \|_2^2 + g(z)$ denotes the proximal operator for the function $g$, and $\alpha_n$ is the step-size for iteration $n$. \revised{\rnum{R2.2}Note that implicitly, we require the regularization functions to have simple proximal operators that can be evaluated efficiently, which is true for most commonly used regularization functions.} Examples of proximal operators include wavelet soft-thresholding for wavelet $\ell 1$ norm, weighting for $\ell 2$ norm and projection for any indicator function for convex sets. We refer the reader to the manuscript of Parikh and Boyd~\cite{Parikh:2013vb} for an overview.

\revised{\rnum{R2.1}Using the CR calculus~\cite{kreutz2009complex}, we can derive exact expressions for the gradient terms.} Substituting the gradient terms, the update steps (\ref{eq:mag_update}, \ref{eq:phase_update}) at iteration $n$ can be explicitly written as:
\begin{equation}
	\begin{aligned}
		r_n &= A^* \left(y - A  (M m_n \cdot e^{\imath  P p }) \right) \\
		m_{n+1} &= \prox_{\alpha_n g_m} \left( m_n +   \alpha_n \text{Re} ( M^*  (e^{-\imath P p }  \cdot r_n ) ) \right)
	\end{aligned}
      \end{equation}
and
 \begin{equation}
	\begin{aligned}
		r_n &= A^* \left(y - A  (M m \cdot e^{\imath  P p_n }) \right) \\
		p_{n+1} &= \prox_{\alpha_n g_p} \left( p_n +  \alpha_n \text{Im}\left( P^*  (M m \cdot e^{-\imath P p_n }  \cdot r_n ) \right) \right)
	\end{aligned}
\end{equation}
where $r_n$ can be interpreted as the residual complex image at iteration $n$.

\begin{figure}[!ht]
  \centering
  \includegraphics[width=\linewidth]{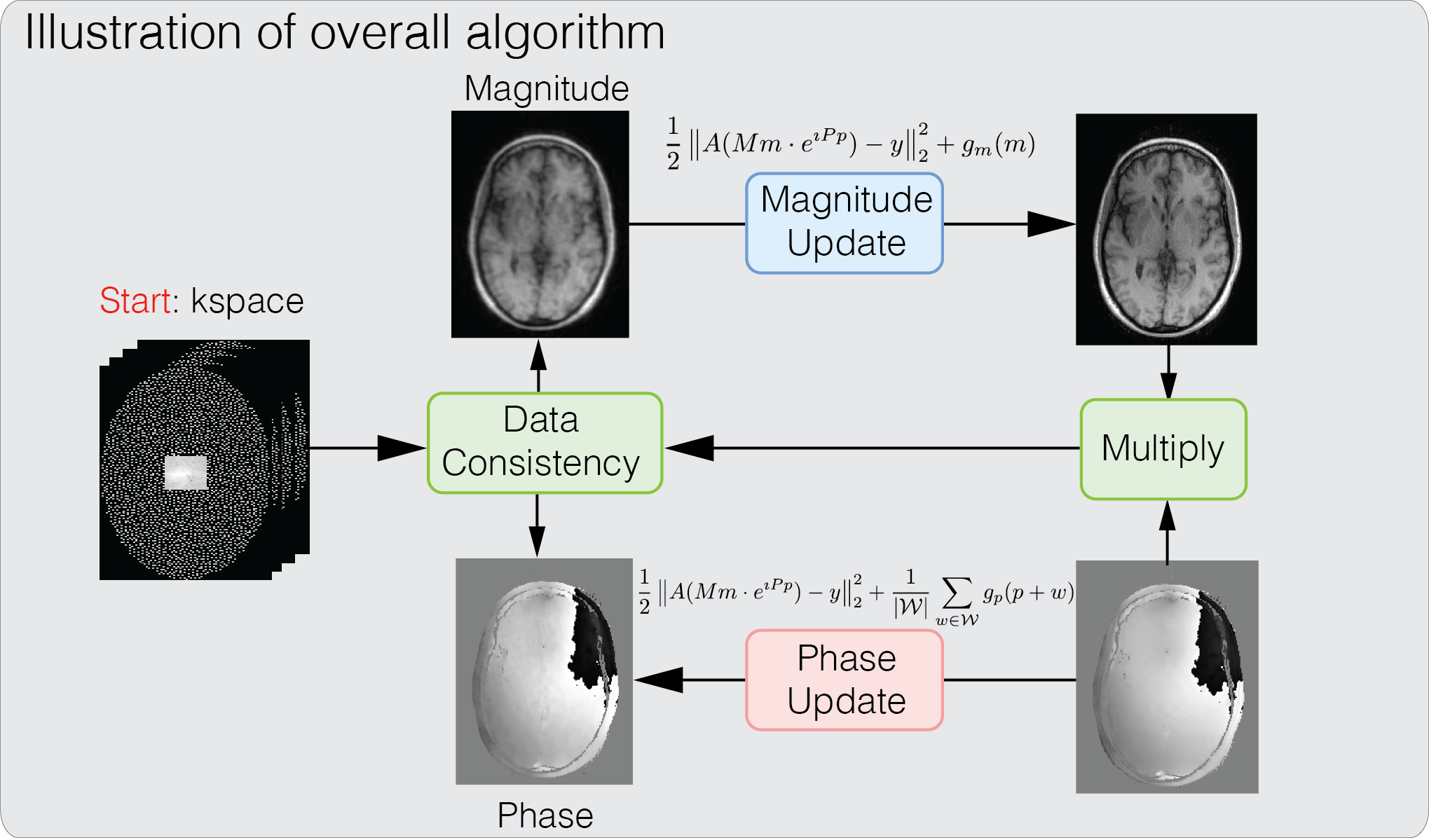}
  \caption{Conceptual illustration of the proposed algorithm. The overall algorithm alternates between magnitude update and phase update. Each sub-problem is solved using proximal gradient descent.}
  \label{fig:alg_overall}
\end{figure}

\begin{figure}[!ht]
  \centering
  \includegraphics[width=\linewidth]{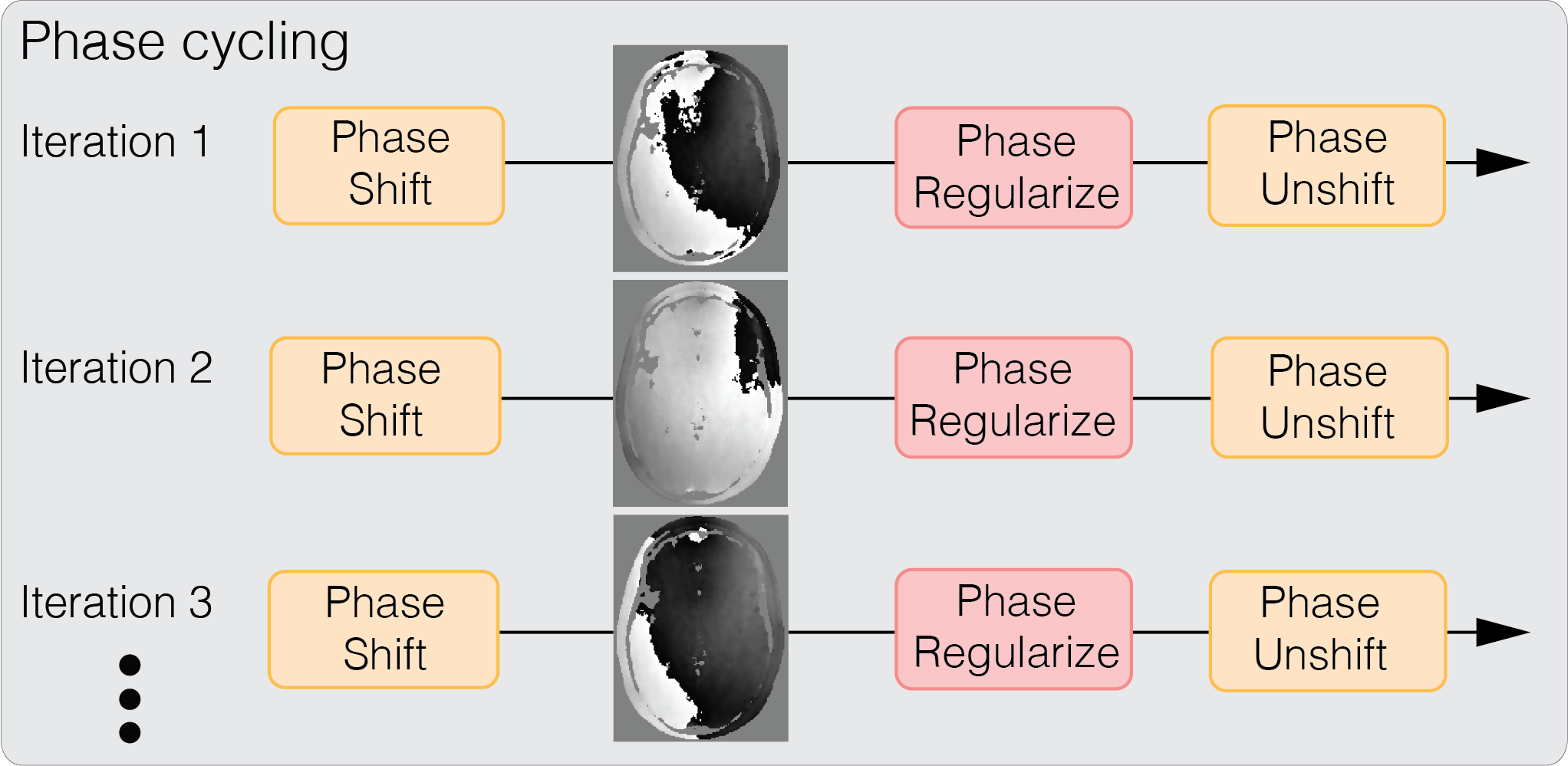}
  \caption{Conceptual illustration of the proposed phase cycling reconstruction technique. Phase cycling achieves robustness towards phase wraps by spreading the artifacts caused by regularizing phase wraps spatially.}
  \label{fig:alg_phase_cycle}
\end{figure}

\subsection*{Phase Cycling}
\label{ssec:phase_cycle}

While the above algorithm converges to a stationary point, this stationary point is very sensitive to the phase wraps in the initial solution in practice. This is because phase wraps are highly discontinuous and are artifacts from the initialization method. Phase regularization in each iteration penalizes these discontinuities, causing errors to accumulate at the same position over iterations, and can result in significant artifacts in the resulting solution. Figure~\ref{fig:phase_error} shows an example of applying smoothing regularization on the phase image by soft-thresholding its Daubechies wavelet coefficients. While the general noise level is reduced, the phase wraps are also falsely smoothened, causing errors around it, as pointed by the red arrows. These errors accumulate over iterations and cause significant artifacts near phase wraps in the reconstructed image, \revised{\rnum{R.12.22}\erased{as shown in figure}as pointed out by the yellow arrow in Figure}~\ref{fig:brain_proposed}. \revised{\rnum{R1.3}The supporting videos~\ref{vid:without_phase_cycling} and~\ref{vid:with_phase_cycling} show the development of the reconstruction results over iterations without and with phase cycling for the experiment in Figure~\ref{fig:brain_proposed}, demonstrating the convergence behavior described above.}

\begin{figure}[!ht]
  \centering
  \includegraphics[width=0.7\linewidth]{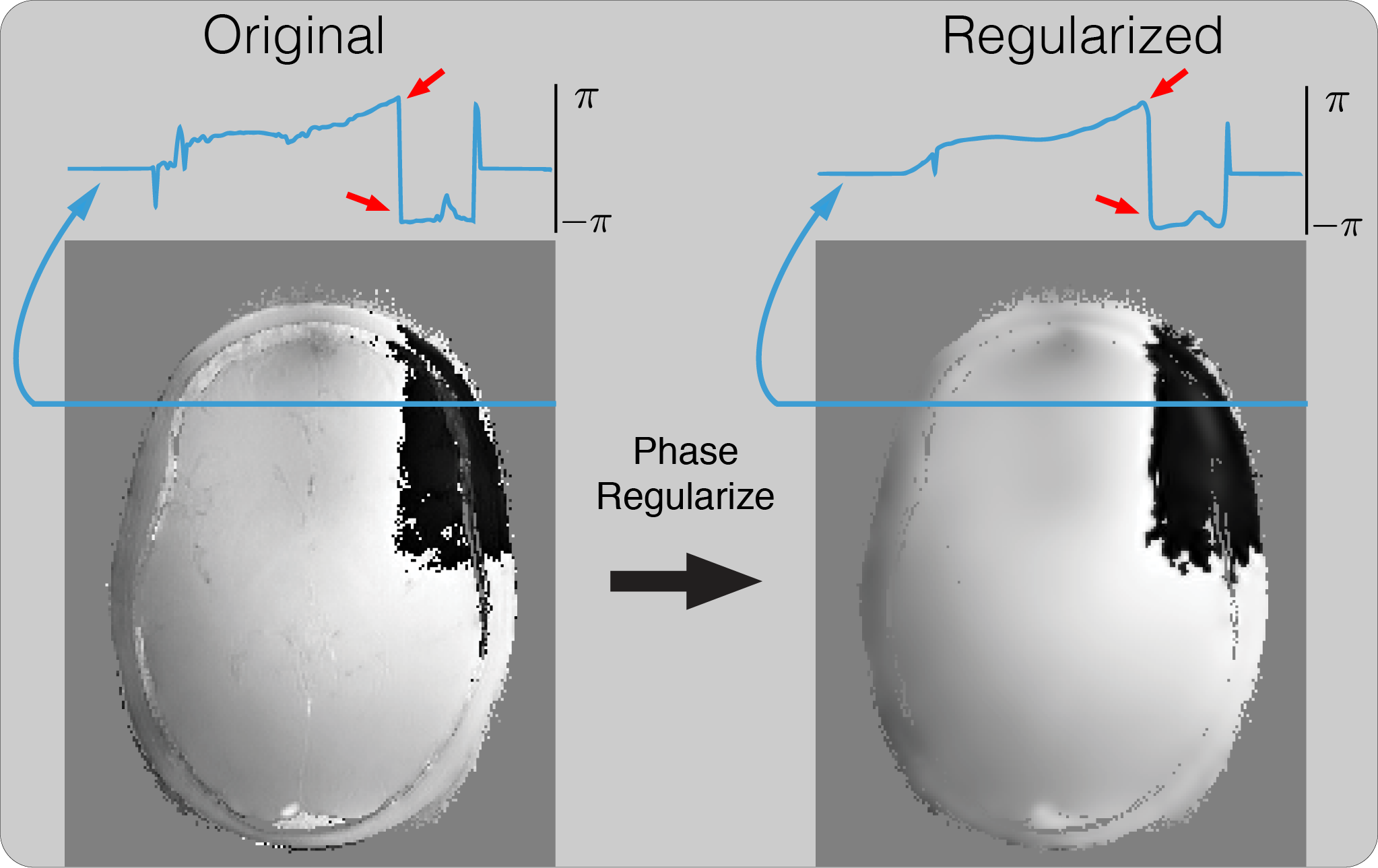}
  \caption{An example of applying smoothing regularization on the phase image by soft-thresholding its Daubechies wavelet coefficients. While the general noise level is reduced, the phase wraps are also falsely smoothened, causing errors around it, as pointed by the red arrow. These errors accumulate over iterations and cause significant artifacts near phase wraps as shown in Figure~\ref{fig:brain_proposed}.}
  \label{fig:phase_error}
\end{figure}

To mitigate these artifacts from regularizing phase wraps, we propose a reconstruction technique, called phase cycling, to make our reconstruction invariant to phase wraps in \revised{\rnum{R2.12.23}the }initial solution. Our key observation is that even though it is difficult to reliably unwrap phase, the position of these phase wraps can easily be shifted to a different spatial location by adding a constant global phase. Hence, artifacts caused by phase regularization can also be shifted spatially to a different location. Our phase cycling method simply proposes to shift the phase wraps randomly over iterations, illustrated in Figure~\ref{fig:alg_phase_cycle}\revised{\rnum{R2.12.24},} to prevent significant error accumulation at the same spatial location. Then over iterations, the artifacts are effectively averaged spatially. 

Concretely, let $\wraps$ be the set of phase wraps generated from the initial solution. Then for each iteration $n$, we randomly draw a phase wrap $w_n$ from $\wraps$ with equal probability, and propose the following phase cycled update for phase images $p$ with fixed magnitude images $m$ at iteration $n$:
\begin{equation}
  \begin{aligned}
    p_{n+1} &= \prox_{\alpha_n g_p} \left( p_n + w_n - \alpha_n \nabla_{p} f(m, p_n) \right) - w_n
  \end{aligned}
\end{equation}
A pseudocode of the proposed algorithm with phase cycling is included in Appendix~\ref{sec:pseudocode}.

Finally, we note that the phase cycled update steps can be viewed as an inexact proximal gradient method applied on the following robust objective function:
\begin{equation}
	\begin{aligned}
		\frac{1}{2} \left\| y - A\left(M m \cdot e^{\imath P p } \right)  \right\|_2^2 +  \vphantom{\frac{1}{2}}g_m ( m ) + \frac{1}{|\wraps|} \sum_{w \in \wraps} g_p (p + w)
	\end{aligned}
	\label{eq:obj_inexact}
      \end{equation}
      where the phase regularization function is averaged over phase wraps. We refer the reader to Appendix~\ref{sec:inexact} for more details.

\section{Methods}
\label{sec:methods} 

The proposed method was evaluated on partial Fourier imaging, water-fat imaging and flow imaging applications. Parallel imaging \revised{\rnum{R3.6.4}\erased{were}was} incorporated in each application. Sensitivity maps were estimated using ESPIRiT~\cite{Uecker:2013hu}, an auto-calibrating parallel imaging method, \revised{\rnum{R2.9}using the Berkeley Advanced Reconstruction Toolbox (BART)~\cite{Uecker:2015fo}}. \revised{\rnum{R2.9}All reconstruction methods were implemented in MATLAB (MathWorks, Natick, MA), and run on a laptop with a 2.4 GHz Intel Core i7 with 4 multi-cores, and 8GB memory.} Unless specified otherwise, the magnitude image was regularized with the $\ell 1$ norm on the Daubechies 4 wavelet transform for the proposed method. \revised{\rnum{R1.2}The number of outer iteration was set to be 100, and the number of inner iteration for both magnitude and phase was set to be 10. The regularization parameters were selected by first fixing the magnitude regularization parameter, and optimizing the phase regularization parameter over a grid of parameters with respect to mean squared error. And then fixing the phase regularization parameter, and optimizing the magnitude regularization parameter similarly. The step-size for the magnitude update was chosen to be $\frac{1}{\lambda_\text{max}(A^* A) \lambda_\text{max}(M^* M)}$, and the step-size for the phase update was chosen to be $\frac{1}{\lambda_\text{max}(A^* A) \lambda_\text{max}(P^* P) \max(|M m|^2)}$, where $\lambda_\text{max}$ denotes the maximum eigenvalue.}

\subsection*{Partial Fourier Imaging}
\label{ssec:method_homodyne}

A fully-sampled dataset from a human knee was acquired on a 3T GE Discovery MR 750 scanner (GE Healthcare, Waukesha, WI) and 8-channel knee coil, with a 3D FSE CUBE sequence, TE/TR = 25/1550 ms, 40 echo train length, image size $320 \times 320 \times 256$, and spatial resolution of $0.5 \times 0.5 \times 0.6\text{ mm}^3$ as described in Epperson et al.~\cite{Epperson:2013vd} and is available online at \url{http://www.mridata.org/}. Another fully-sampled dataset from a human brain was acquired on 1.5T GE Signa scanner (GE Healthcare, Waukesha, WI) with a $8$-channel head coil, 3D GRE sequence, TE/TR = 5.2 ms / 12.2 ms, image size of $256 \times 256 \times 230$ and spatial resolution of 1 mm. 2D slices were extracted along the readout direction for the experiments. \revised{Image masks for displaying the phase images were created by thresholding the bottom 10\% of the magnitude images.}

A partial Fourier factor of $5/8$ was retrospectively applied on both datasets. The brain dataset was further retrospectively under-sampled by $4$ with variable density Poisson-disk pattern and a $24 \times 24$ calibration region. The proposed method with and without phase cycling were applied on the under-sampled datasets and compared. $\ell 1$ regularization was imposed on the Daubechies 6 wavelet domain for the phase image. \revised{\rnum{R2.12.25}\erased{Homodyne}The homodyne} method from Bydder et al.~\cite{Bydder:2005bya} with $\ell1$ regularization on the wavelet domain was applied and compared. \revised{\rnum{R2.3}The method was chosen for comparison because it was shown to be robust to errors in the phase estimate as it penalizes the imaginary component instead of enforcing the image to be strictly real.} \revised{\rnum{R1.4}The original formulation included only $\ell 2$ regularization on the real and imaginary components separately. To enable a fair comparison using similar image sparsity models, we modified the method to impose $\ell 1$ wavelet regularization on the real and imaginary components separately to exploit wavelet sparsity, which achieved strictly smaller mean squared error than the original method.} \revised{\rnum{R1.2}The number of iterations was set to be 1000. The regularization parameters were set similarly to how the magnitude and phase regularization parameters were selected.}

\revised{\rnum{R3.2}Besides visual comparison, the quality of the reconstructed magnitude images was evaluated using the peak signal-to-noise ratio (PSNR). Given a reference image $x_\text{ref}$, and a reconstructed image $x_\text{rec}$, PSNR is defined as:}
\begin{equation}
	\begin{aligned}
          \text{PSNR}(x_\text{ref}, x_\text{rec}) = 20 \log_{10}\left( \frac{\max(x_\text{ref})}{\| x_\text{ref} - x_\text{rec} \|_2} \right)
	\end{aligned}
\end{equation}

For our proposed method, the magnitude image and phase image were initialized from the zero-filled reconstructed image, that is $m = | A^\top y |$ and $p = \angle(A^\top y)$. 

\subsection*{Water-Fat Imaging}
\label{ssec:method_waterfat}

Fully sampled water-fat datasets were obtained from the ISMRM Water-Fat workshop toolbox (available online at \url{http://www.ismrm.org/workshops/FatWater12/data.htm}). In particular, an axial slice of the liver with 8-channel coil array dataset was used, which also appeared in the paper of Sharma et al.~\cite{Sharma:2011ca}. The dataset was acquired on a 3T Signa EXCITE HDx system (GE Healthcare, Waukesha, WI), using a GE-investigational IDEAL 3D spoiled-gradient-echo sequence at three TE points with $TE= [2.184, 2.978, 3.772]$ ms, $BW=\pm 125$ kHz, flip angle of $5$ \revised{\rnum{R2.12.26}\erased{degree}degrees} and a $256 \times 256$ sampling matrix. \revised{Image masks for displaying the phase images were created by thresholding the bottom 10\% of the root-sum-of-squared of the magnitude images.}

The liver dataset was retrospectively under-sampled by 4 with a variable density Poisson-Disk sampling pattern. Our proposed method was applied and compared with algorithm of Sharma et al.~\cite{Sharma:2011ca} both for the fully-sampled and under-sampled datasets.  \revised{\rnum{R1.2}The implementation in the Water-Fat workshop toolbox was used with modification to impose the same wavelet transforms as the proposed method, and the default parameters were used.}

An axial slice of the thigh dataset from the ISMRM Water-Fat workshop toolbox was also used, with the same parameters. The dataset was retrospectively under-sampled by 4 with a variable density Poisson Disk sampling pattern and an \revised{\rnum{R1.8}\erased{addition}additional} 9/16 partial Fourier factor. Our proposed method was applied and compared with the result of applying the algorithm of Sharma et al. on the fully-sampled dataset. An $\ell 1$ regularization on the Daubechies-4 wavelet transform of the image phase was applied.

For our proposed method, the field map was initialized as zero images, the magnitude images were initialized as $(m_\text{water}, m_\text{fat})^\top = | M^\top A^\top y |$, and the phase images $(p_\text{water}, p_\text{fat})^\top$ were extracted from $\angle( M^\top A^\top y)$.

\subsection*{Divergence-free Regularized Flow Imaging}
\label{ssec:method_flow}

Four 4D flow datasets of pediatric patients with tetrahedral flow encoding were acquired with 20 cardiac phases, 140 slices and \revised{\rnum{R2.12.27}\erased{a}an} average spatial resolution of $1.2 \times 1.2 \times 1.5 \text{mm}^3$. The 4D flow acquisitions were performed on a 1.5T Signa Scanner (GE Healthcare, Waukesha, WI) with an 8 channel array using a spoiled gradient-echo-based sequence. The acquisitions were prospectively under-sampled by \revised{\rnum{R2.12.28}\erased{about 4}an average factor of 3.82} with variable density Poisson-disk undersampling. The flip angle was 15 degrees and average TR/TE was 4.94/1.91 ms. The performance of the proposed method was compared with $\ell 1$-ESPIRiT~\cite{Uecker:2013hu}, a PI + CS algorithm. Volumetric eddy-current correction was performed on velocity data. Segmentations for flow calculations were done manually on the aorta (AA) and pulmonary trunk (PT). Net flow rate and regurgitant fraction were calculated for each segmentation. \revised{\rnum{R2.5}Since the datasets did not contain phase wraps in the region of interest, phase unwrapping was not performed.} \revised{Image masks for displaying the phase images were created by thresholding the bottom 20\% of the magnitude images.} \revised{\rnum{R3.5}One of the $\ell 1$ ESPIRiT reconstruction result was further processed with divergence-free wavelet denoising~\cite{Ong:2014gq} to compare with the proposed method.}

Another flow dataset of pediatric patient, with randomized flow encode undersampling using the VDRad sampling pattern~\cite{Cheng:2015jl} and a partial readout factor of 0.7, was acquired with 20 cardiac phase. The proposed method was applied and compared to $\ell 1$-ESPIRiT.

For our proposed method, an $\ell 1$ regularization on the divergence-free wavelet transform~\cite{Ong:2014gq} of the flow images was used to impose divergence-free constraint and an $\ell 1$ regularization on the Daubechies 4 wavelet coefficients of the \revised{\erased{reference}background} phase was used to impose smoothness. The flow images were initialized as zero images and the \revised{\erased{reference}background} phase image $p_\text{bg}$ was extracted from the first velocity encode of $\angle( A^\top y)$.


\section{Results}
\label{sec:results}

In the spirit of reproducible research, we provide a software package in MATLAB to reproduce most of the results described in this paper. The software package can be downloaded from:

\url{https://github.com/mikgroup/phase_cycling.git}

\subsection*{Partial Fourier Imaging}
\label{ssec:result_homodyne}

Supporting Figure~\ref{fig:knee} shows the partial Fourier reconstruction results combined with PI on the knee dataset. The figure compares the proposed reconstruction with and without phase cycling along with the homodyne reconstruction method described in Bydder et al.~\cite{Bydder:2005bya}. For the reconstruction without phase cycling, significant artifacts near the phase wraps can be seen in the magnitude image, as pointed by the yellow arrow. The proposed reconstruction with phase cycling also performs comparably with the state-of-the-art and did not display significant artifacts. \revised{\rnum{R3.2}In terms of PSNR, the method of Bydder et al. resulted in 34.55 dB, the proposed method without phase cycling resulted in 32.83 dB, and the proposed method with phase cycling resulted in 34.93 dB.} \revised{\rnum{R2.9}One instance of our Matlab implementation of the proposed method took 4 minutes and 6 seconds.}

Figure~\ref{fig:brain_proposed} shows the partial Fourier reconstruction results combined with PI and CS on the brain dataset with the proposed method. The figure compares the proposed reconstruction with and without phase cycling. Again, without phase cycling, significant artifacts can be seen in the magnitude image near phase wraps in the initial solution, pointed by the yellow arrow. Reconstruction with phase cycling does not display these artifacts. \revised{Figure~\ref{fig:brain_compare} shows the results for the method of Bydder et al. and Zhao et al. As pointed out by the red arrows, the method of Bydder et al. shows higher error in the magnitude image compared to proposed method with phase cycling in these regions. The method of Zhao et al. shows higher magnitude image error in general, and displays staircase artifacts in the phase image, which are common in total variation regularized images, as pointed by the yellow arrow.} \revised{\rnum{R3.2}In terms of PSNR, the method of Bydder et al. resulted in 33.64 dB, the method of Zhao et al. resulted in 30.62 dB, the proposed method without phase cycling resulted in 30.35 dB, and the proposed method with phase cycling resulted in 34.91 dB.} \revised{\rnum{R2.9}One instance of our Matlab implementation of the proposed method took 1 minute and 53 seconds.} \revised{\rnum{R1.1}We note that the severity of the artifact in the reconstruction without phase cycling for the brain dataset is much stronger than for the knee dataset because a higher regularization was needed to obtain lower reconstruction mean squared error. This is because the brain dataset was further under-sampled for CS. Larger regularization led to larger thresholding errors around the phase wraps and hence more significant artifacts in the resulting reconstructed images.}

\begin{figure}[!ht]
\begin{center}
\includegraphics[width=\linewidth]{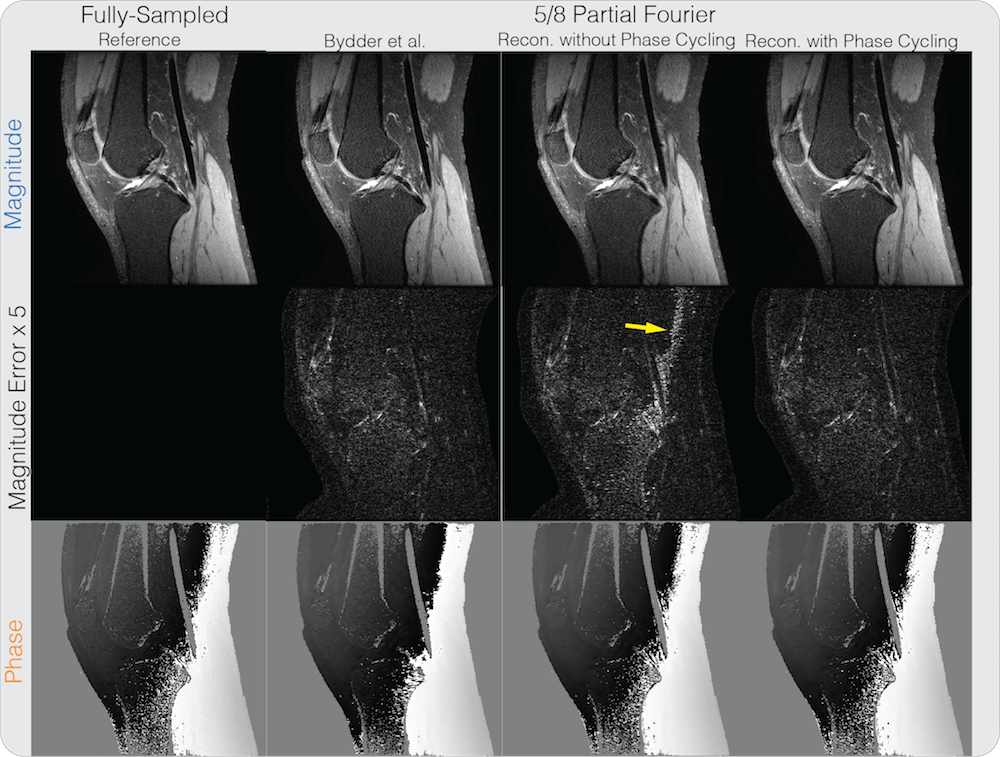}
\caption{Partial Fourier + PI reconstruction results on a knee dataset. Without phase cycling, significant artifacts can be seen in the magnitude image near phase wraps in the initial solution, pointed by the yellow arrow. With phase cycling, these artifacts were reduced and the result is comparable to the robust iterative partial Fourier method with $\ell 1$ wavelet described in Bydder et al.}
\label{fig:knee}
\end{center}
\end{figure}

\begin{figure}[!ht]
\begin{center}
\includegraphics[width=0.8\linewidth]{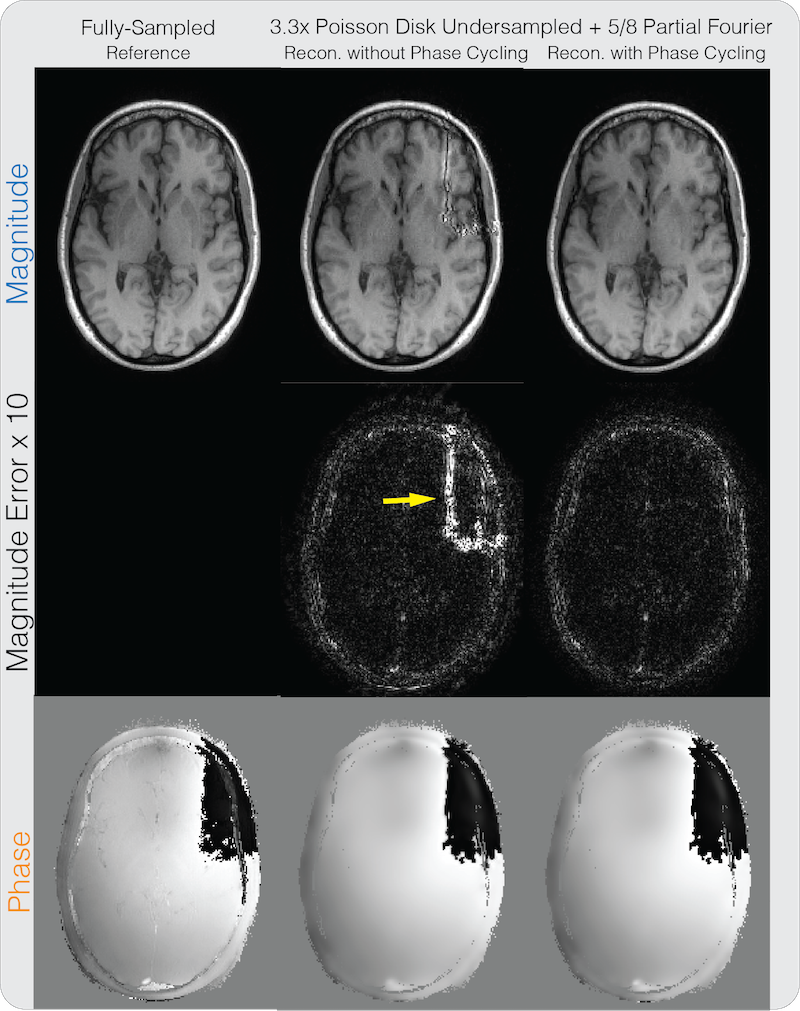}
\caption{Partial Fourier + PI + CS reconstruction results on a brain dataset for the proposed method. Similar to Supplementary Figure~\ref{fig:knee}, without phase cycling, siginificant artifacts can be seen in the magnitude image near phase wraps in the initial solution, as pointed by the yellow arrow.}
\label{fig:brain_proposed}
\end{center}
\end{figure}

\begin{figure}[!ht]
\begin{center}
\includegraphics[width=0.8\linewidth]{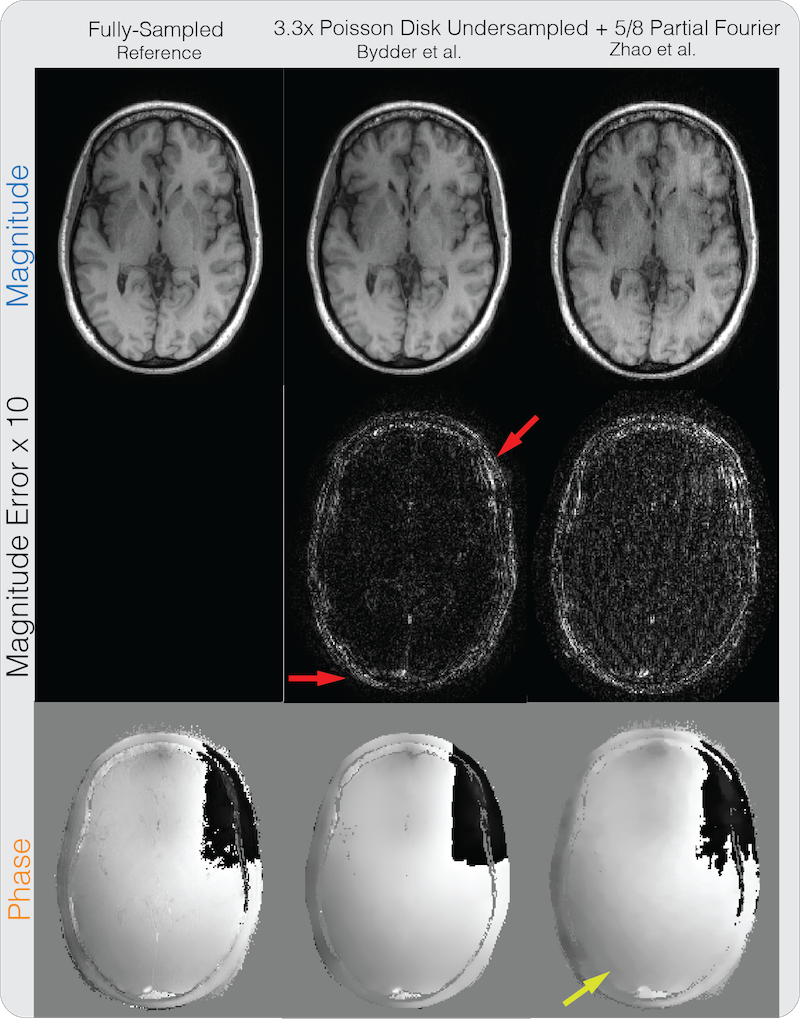}
\caption{Partial Fourier + PI + CS reconstruction comparison results on the same brain dataset in Figure~\ref{fig:brain_proposed}. The method of Bydder shows higher error in the magnitude image compared to proposed method with phase cycling in regions pointed out by the red arrows. The method of Zhao et al. shows higher magnitude image error in general, and displays staircase artifacts in the phase image, which are common in total variation regularized images, as pointed by the yellow arrow.}
\label{fig:brain_compare}
\end{center}
\end{figure}

\subsection*{Water-Fat Imaging}
\label{ssec:result_waterfat}

Supporting Figure~\ref{fig:liver} shows the water-fat reconstruction results on the liver dataset, combined with PI and CS for the under-sampled case. For the fully sampled case, the water and fat images from the proposed method with phase cycling \revised{\rnum{R2.12.29}\erased{is comparable with}are comparable with the} state-of-the-art water-fat reconstruction result using the method of Sharma et al.~\cite{Sharma:2011ca}. Reconstruction from under-sampled data with the proposed method also results in similar image quality as the fully-sampled data and is consistent with the result shown in Sharma et al.~\cite{Sharma:2011ca}. \revised{\rnum{R2.9}One instance of our Matlab implementation of the proposed method took 8 minutes and 27 seconds.}

\begin{figure}[!ht]
\begin{center}
\includegraphics[width=\linewidth]{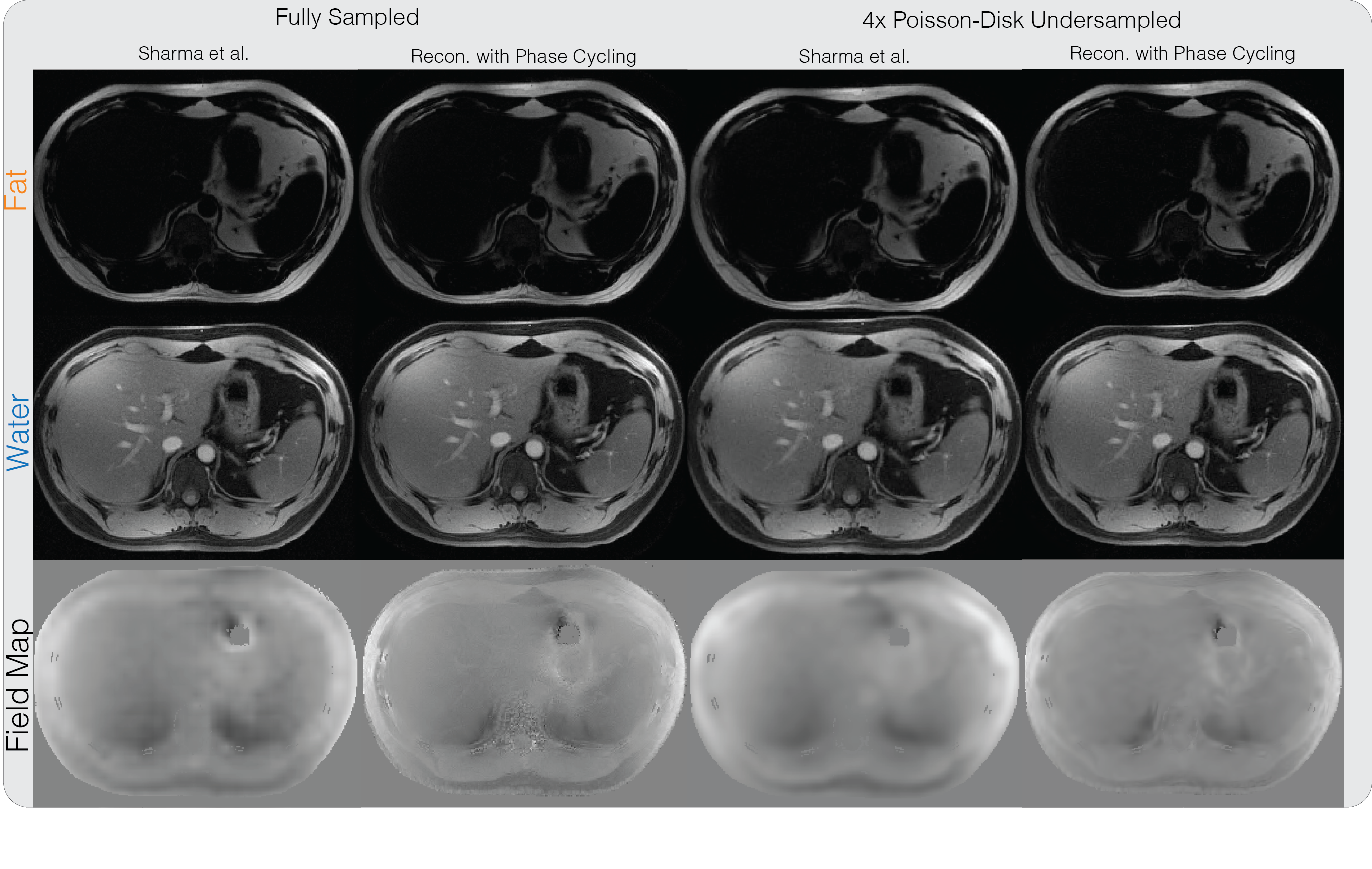}
\caption{Water-fat + PI + CS reconstruction result on a liver dataset with three echoes. Both the method from Sharma et al. and our proposed method produce similar water and fat images on the fully-sampled dataset and the retrospectively under-sampled dataset.}
\label{fig:liver}
\end{center}
\end{figure}

Figure~\ref{fig:thigh} shows the water-fat reconstruction results on the thigh dataset, combined with partial Fourier, PI and CS. Our proposed method produces similar water and fat images on a partial Fourier dataset, as the fully-sampled reconstruction using the method of Sharma et al.~\cite{Sharma:2011ca}. This demonstrates the feasibility of performing joint partial Fourier and water fat image reconstruction along with PI and CS using the proposed method. \revised{\rnum{R2.9}One instance of our Matlab implementation of the proposed method took 8 minutes and 23 seconds.}

\begin{figure}[!ht]
\begin{center}
\includegraphics[width=\linewidth]{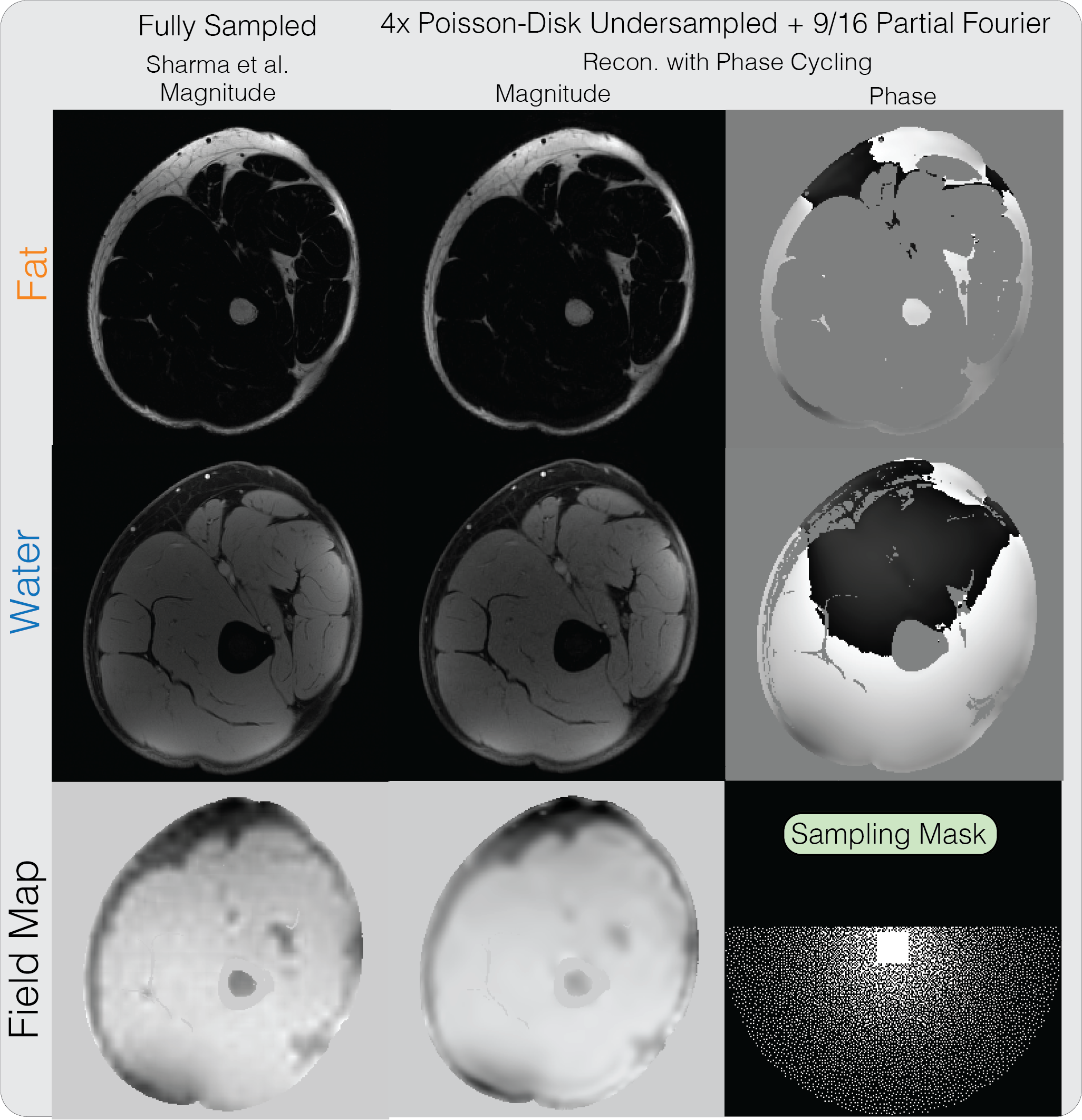}
\caption{Water-fat + partial Fourier + PI + CS reconstruction result on a thigh dataset. Our proposed method produce similar water and fat images on a under-sampled partial Fourier dataset compared to the fully-sampled reconstruction using the method from Sharma et al.}
\label{fig:thigh}
\end{center}
\end{figure}

\subsection*{Divergence-free Regularized Flow Imaging}
\label{ssec:result_4dflow}

Figure~\ref{fig:dfw_quant} shows the net flow rates calculated for four patient data. Both $\ell 1$-ESPIRiT reconstruction and proposed reconstruction resulted in similar flow rates. Maximum difference in regurgitant fractions was 2\%. The comparison showed that in these four cases, our proposed method provided comparable measurements to those obtained from $\ell 1$-ESPIRiT, which were shown to correlate with 2D phase contrast flow rates~\cite{Hsiao:2012kx}. \revised{\rnum{R1.6}\erased{Study 2 results are}A representative velocity image and speed map are} shown in \revised{\rnum{R2.12.30}\erased{figure}Figure}~\ref{fig:dfw}. Visually, the proposed method reduces incoherent artifacts and noise \revised{compared to the other results, especially in regions pointed by the red arrows where there should be no fluid flow.\erased{while providing similar magnitude images.}}

Figure~\ref{fig:dfw_pf} shows the result of reconstructing a partial readout 4D flow dataset with randomized velocity encoding. The reconstructed magnitude image has significantly reduced blurring from partial readout, compared to the $\ell 1$-ESPIRiT result. We also note that the velocity images are not masked and that velocities in low magnitude regions are naturally suppressed with the proposed reconstruction. \revised{\rnum{R2.9}One instance of our Matlab implementation of the proposed method took on the order of three hours.}

\begin{figure}[!ht]
  \begin{center}
    \includegraphics[width=0.5\linewidth]{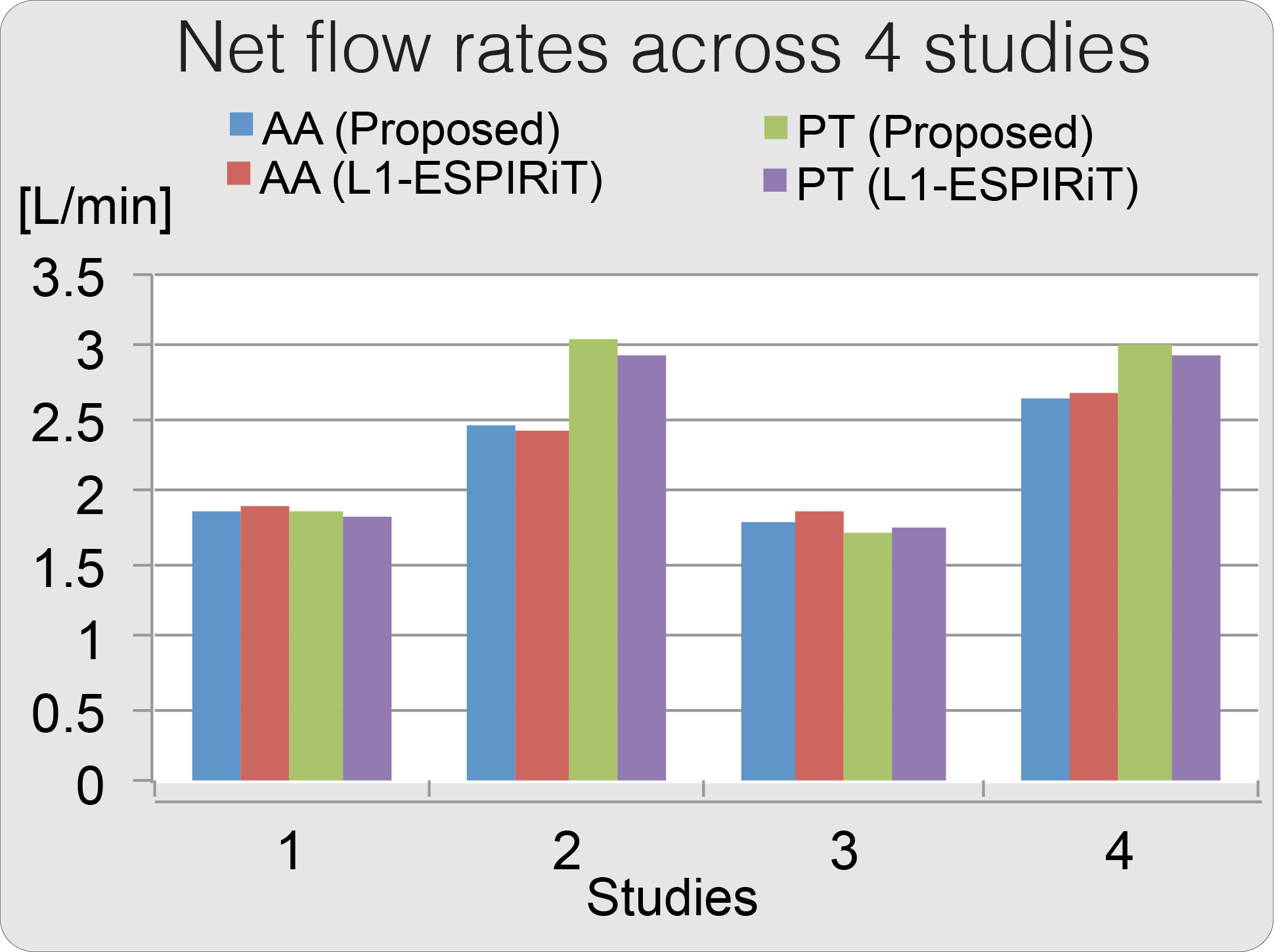}
    \caption{Net flow rates across 4 studies for the proposed method compared with $\ell 1$-ESPIRiT, calculated across segmentations of aorta (AA) and pulmonary trunk (PT). Both $\ell 1$-ESPIRiT reconstruction and proposed reconstruction result in similar flow rates.}
    \label{fig:dfw_quant}
  \end{center}
\end{figure}

\begin{figure}[!ht]
\begin{center}
\includegraphics[width=\linewidth]{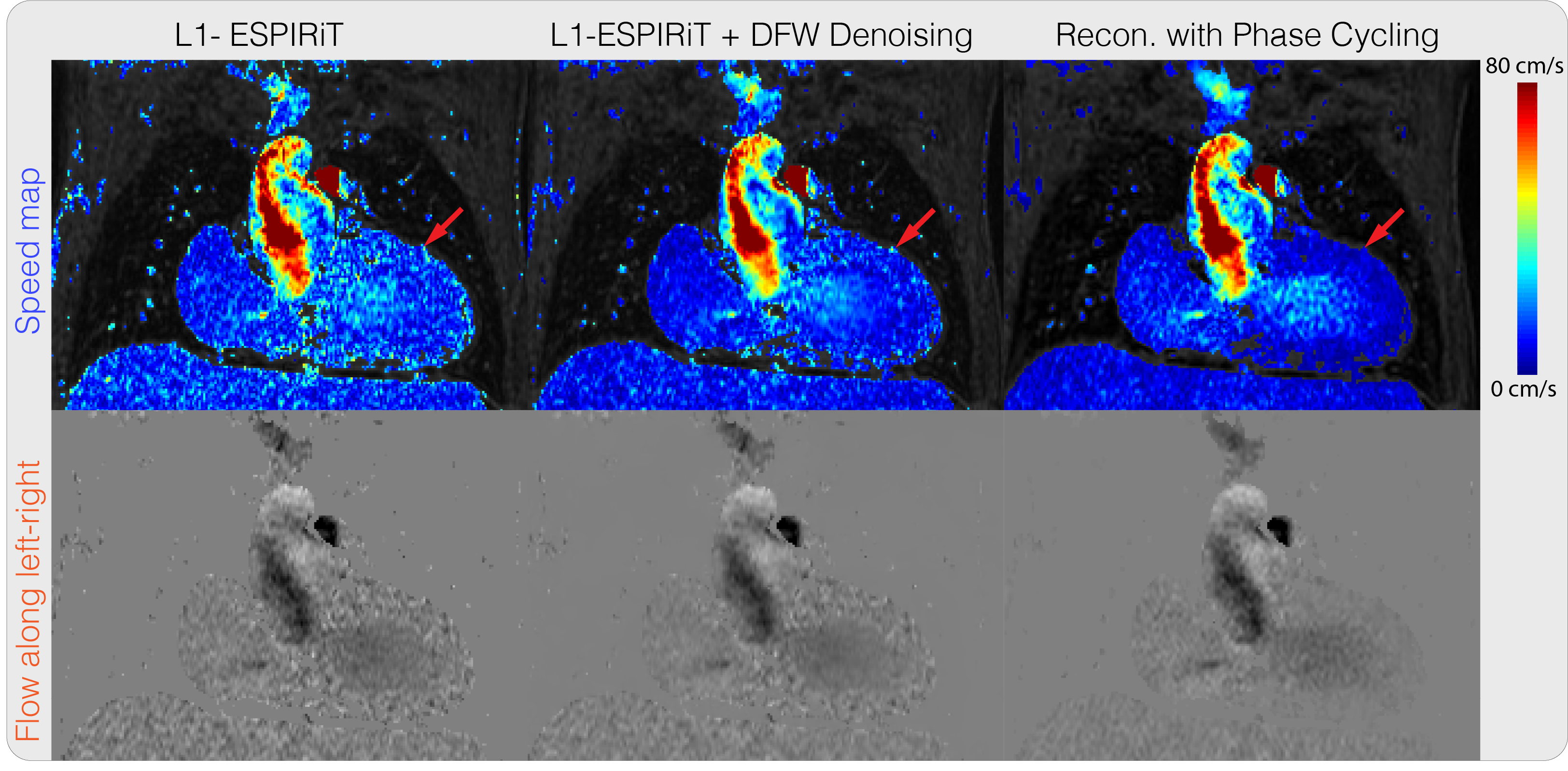}
\caption{Divergence-free wavelet regularized flow imaging + PI + CS reconstruction result on a 4D flow dataset, compared with $\ell 1$-ESPIRiT and $\ell 1$-ESPIRiT followed by divergence-free wavelet denoising. Visually, the proposed method reduces incoherent artifacts and noise compared to the other results, especially in regions pointed by the red arrows where there should be no fluid flow.}
\label{fig:dfw}
\end{center}
\end{figure}

\begin{figure}[!ht]
\begin{center}
\includegraphics[width=\linewidth]{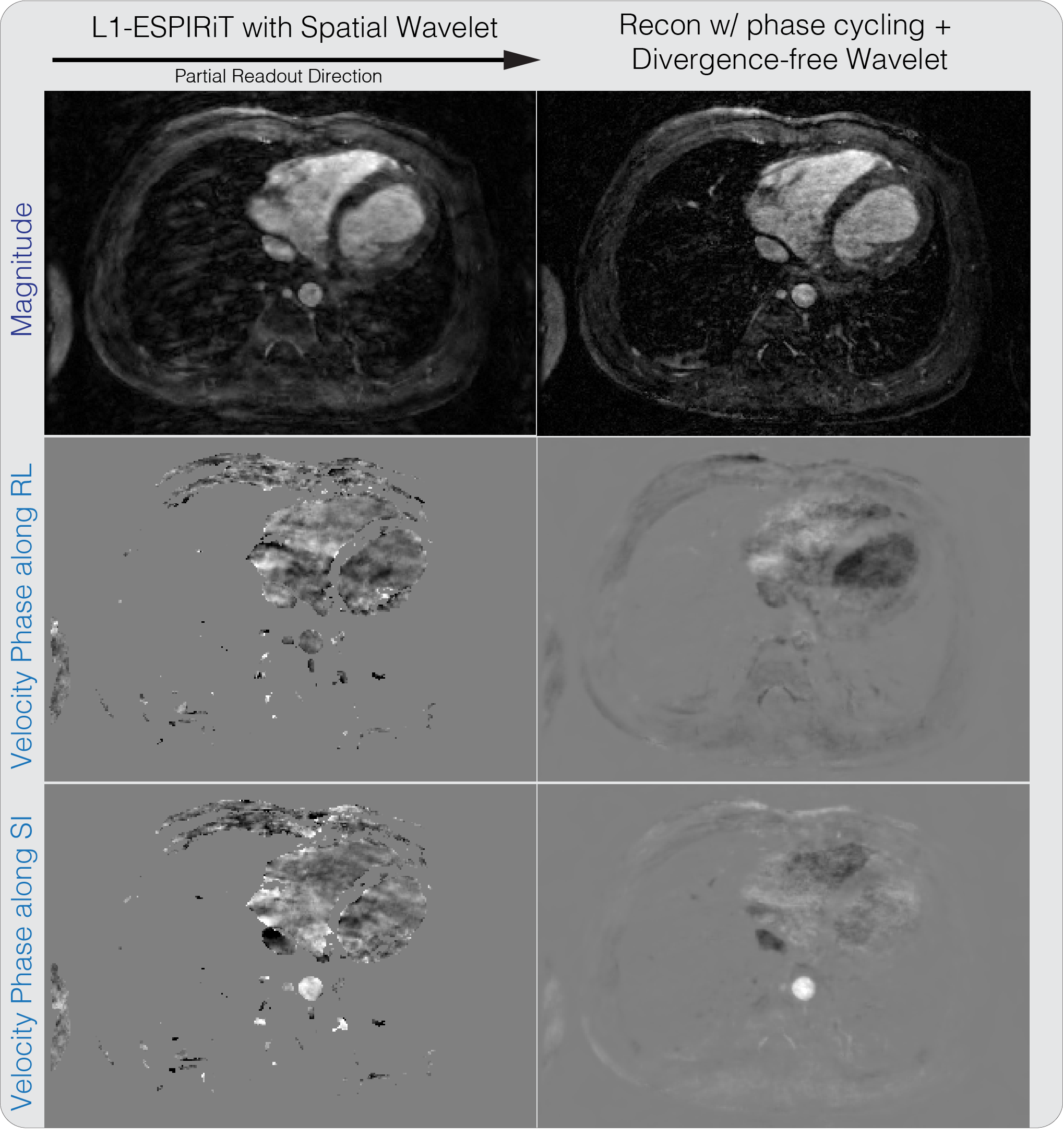}
\caption{Divergence-free wavelet regularized flow imaging + PI + CS reconstruction results on a 4D flow dataset. The reconstructed magnitude image has significantly reduced blurring from partial readout, compared to the $\ell 1$-ESPIRiT result.}
\label{fig:dfw_pf}
\end{center}
\end{figure}


\section*{Discussion}
\label{sec:discussion}

In this work, we describe a unified framework and algorithm for phase regularized reconstructions. By presenting various phase regularized image reconstructions within the same framework, we are able to combine these phase sensitive imaging methods to improve the reconstructed results\revised{\erased{ and perform joint reconstruction}}: \revised{\erased{our results in partial Fourier + water-fat imaging + PI + CS, and divergence-free regularization + partial Fourier + PI + CS show significant improvements in resolution and image quality via joint optimization.}Our result in partial Fourier + water-fat imaging + PI + CS shows the ability to further accelerate water-fat imaging with PI + CS by combining it with partial Fourier, while achieving comparable image quality as the fully sampled images. Our result in divergence-free flow + partial Fourier + PI + CS also shows the ability to achieve improvements in apparent image resolution over standard PI + CS reconstruction.} The proposed framework shows \revised{\rnum{R2.12.31}\erased{promises}promise} and encourages the use of joint application of phase sensitive imaging techniques.

\revised{One advantage of our proposed method is that more appropriate phase regularization can be enforced for each application. In particular, we compared our proposed method with the methods of Bydder et al., and Zhao et al. for partial Fourier imaging. The method of Bydder et al. requires imposing regularization on the image imaginary component, which does not necessarily correlate with regularization of the phase image. The method of Zhao et al., on the other hand, only supports finite difference penalties, which can result in staircase artifacts, shown in Figure~\ref{fig:brain_compare}. Our proposed method can impose a more general class of phase regularization, and in the case of partial Fourier imaging, the proposed method with Daubechies-6 wavelet sparsity constraint on the phase image resulted in the highest PSNR among the compared methods. For water-fat imaging, we compared our proposed method with the method of Sharma et al., and achieved similar image quality for both fully-sampled and under-sampled datasets. With phase cycling, our proposed method can be extended to include partial Fourier imaging in a straightforward way. Since the method of Sharma et al. solves for the absolute phase, it is unclear whether their proposed restricted subspace model can be extended to enforce smoothness of water and fat phase images, which often contain phase wraps, and more effort is needed to investigate such extension. Finally, through comparing the proposed method on divergence-free flow imaging and $l1$-ESPIRiT with divergence-free wavelet denoising, we have shown the advantage of joint PI + CS reconstruction utilizing phase structure over separate application of PI + CS reconstruction followed by phase image denoising.}

Further improvement in these applications can be obtained by incorporating additional system information. \revised{\erased{For example, multi-peak model can be used in water-fat image reconstruction, and in 4D flow reconstruction, temporal constraints can be added.}For example, multi-peak model can be used in water-fat image reconstruction by extending the forward model to become:}
\begin{equation}
  \begin{aligned}
    y_e &= F_e  S \left( \left(m_\text{water} e^{\imath  p_\text{water}} + m_\text{fat} e^{\imath  p_\text{fat}} \left(
          \sum_{j=1}^J a_j e^{\imath  t_e 2 \pi \Delta f_j} \right) \right) e^{\imath  t_e p_\text{field}} \right) + \eta_e &\text{for } e = 1, \hdots, E
  \end{aligned}
\end{equation}
\revised{where $J$, $a_j$ and $\Delta f_j$ are the number of fat peaks, the relative amplitude for the $j$th peak, and the frequency difference between the $j$th fat peak and water respectively. Temporal constraints can also be added in 4D flow reconstruction to improve the result. One example would be separately enforcing group sparsity on the Daubechies wavelet coefficients of the magnitude images over time, and divergence-free wavelet coefficients of velocity phase images over time, via the group $\ell 1$ norm.}

\revised{
  Similar to other iterative methods, our proposed method requires parameter selections for the regularization functions. One disadvantage of our proposed method over standard PI + CS methods is that we have two parameters to tune for, and thus requires more effort in selecting these parameters. On the other hand, since phase values are bounded between $\pi$ and $-\pi$, we found that the phase regularization parameter often translates fairly well between experiments with similar settings. That is, a fixed phase regularization parameter results in similar smoothness for different scans with similar undersampling factor and noise level.
}

Since we aim for the joint reconstruction of phase images, phase unwrapping becomes a difficult task. The proposed phase cycling provides an alternative and complementary way of regularizing phases during reconstruction, without the need for phase unwrapping. In cases when the absolute phase is needed, for example in flow imaging \revised{\rnum{R2.5}with low velocity encodes}, phase unwrapping can be performed on the phase regularized reconstructed images, which has the potential to improve phase unwrapping performance due to the reduced artifacts.

Finally, we note that the proposed method with phase cycling still requires a good phase initialization, as the overall objective is non-convex. Our proposed phase cycling only enables the ability to regularize phase without phase unwrapping. In particular, if the initial phase estimate is randomly initialized, then the resulting reconstruction will be much worse than the one with a good initialization. Finding a good phase initialization for all phase imaging methods remains an open problem. Phase initialization for our proposed method is still manually chosen depending on the application, as described in the Methods section. In our \revised{\rnum{R2.12.32}\erased{experiements}experiments}, we found that often either initializing as the phase of the zero-filled reconstructed image, or simply a zero-valued phase image provides good initialization for our proposed method. \revised{\rnum{R3.4}We note that a better initial solution using existing PI + CS methods, such as $\ell 1$-ESPIRiT, can potentially result in a more accurate solution for our proposed method. In our experiments, we found that zero-filled images as starting solutions have already provided adequate starting points with much lower computational cost, and initializing with $\ell 1$ ESPIRiT did not provide noticeable improvements. However, for more aggressively accelerated scans, we expect a better initialization can improve the reconstructed result.}

\section*{Conclusion}
\label{sec:conclusion}

The proposed phase cycling reconstruction provides an alternative way to perform phase regularized reconstruction, without the need of performing phase unwrapping. The proposed method showed reduction of artifacts compared to reconstructions without phase cycling. \revised{The advantage of supporting arbitrary regularization functions, and general magnitude and phase linear models was demonstrated by comparing with other state-of-the-art methods.}\revised{\erased{Joint reconstruction of various combination of partial Fourier, water-fat and divergence-free regularized flow reconstruction \revised{\rnum{R1.8}\erased{were}was} demonstrated, and showed improvements in resolution and quality over separate reconstructions.} Joint reconstruction of partial Fourier + water-fat imaging + PI + CS, and partial Fourier + divergence-free regularized flow imaging + PI + CS were demonstrated.} The proposed method unifies reconstruction of phase sensitive imaging methods and encourages their joint application.


\begin{appendix}
  \section{Phase imaging representation under the general forward model}
  \label{sec:water_fat_flow_general}

 \revised{\rnum{R2.12.17}We first consider representing partial Fourier imaging under the general forward model. Let us define $I$ to be the identity operator with input and output size equal to the image spatial size. We also define the magnitude, phase and complex linear operators to be $M = I$, $P = I$, and $A = F S$. Then partial Fourier imaging forward model can be represented as $y = A(Mm \cdot e^{\imath Pp}) + \eta$.}

  \revised{Next, to represent water-fat imaging under the general forward model, let us define the magnitude images $m$ to be $(m_\text{water}, m_\text{fat})^\top$, and phase images to be $(p_\text{water}, p_\text{fat}, p_\text{field})^\top$. We also define the magnitude operator $M$, phase operator $P$, and the complex operator $A$ to be:}
  \begin{equation}
    \begin{aligned}
      M &=
      \begin{pmatrix}
        I & 0\\
        0 & I\\
        \multicolumn{2}{c}{$\vdots$}\\
        I & 0\\
        0 & I\\
      \end{pmatrix}~~
      P &=
      \begin{pmatrix}
        I & 0 & t_1\\
        0 & I & t_1\\
        \multicolumn{3}{c}{$\vdots$}\\
        I & 0 & t_E\\
        0 & I & t_E\\
      \end{pmatrix}~~
      A &=
      \begin{pmatrix}
        F_1 S & F_1 S e^{j t_1 2 \pi \Delta f}& \hdots & 0 & 0\\
        && \ddots && \\
        0 & 0 & \hdots & F_E S & F_E S e^{j t_E 2 \pi \Delta f}\\
      \end{pmatrix} \\
    \end{aligned}
  \end{equation}
  \revised{where the $2 \times 2$ identity block in $M$ is repeated $E$ times. Then the water-fat imaging forward model~\eqref{eq:cs_d} can be represented as $y = A(Mm \cdot e^{\imath Pp}) + \eta$.}

  \revised{Finally we consider representing flow imaging under the general forward model. For concreteness, we consider the four point balanced velocity encoding as an example. Recall that we define $p = (p_{bg}, p_x, p_y, p_z)^\top$. Let us define the magnitude operator $M$, phase operator $P$, and the complex operator $A$ to be:}
  \begin{equation}
    \begin{aligned}
      M &=
      \begin{pmatrix}
        I\\
        I\\
        I\\
        I\\
      \end{pmatrix}~~
      P &=
      \begin{pmatrix}
        I & - I & - I & -I \\
        I & + I & + I & -I \\
        I & + I & - I & +I \\
        I & - I & + I & +I \\
      \end{pmatrix}~~
      A &=
      \begin{pmatrix}
        F_1 S & 0 & 0 & 0\\
        0 & F_2 S & 0 & 0\\
        0 & 0 & F_3 S & 0\\
        0 & 0 & 0 & F_4 S\\
      \end{pmatrix} \\
    \end{aligned}
  \end{equation}
  \revised{Then the flow imaging forward model~\eqref{eq:flow_d} can be represented as $y = A(Mm \cdot e^{\imath Pp}) + \eta$.}
  
  \section{Pseudocode for phase regularized reconstruction with phase cycling}
  \label{sec:pseudocode}

  Algorithm~\ref{alg:pseudocode} summarizes the proposed reconstruction method with phase cycling.

\begin{algorithm}
  \caption{Pseudocode for phase regularized reconstruction with phase cycling}
  \label{alg:pseudocode}
  \textbf{Input:}\\
  $y$ - observed k-space\\
  $m_0$ - initial magnitude images\\
  $p_0$ - initial phase images\\
  $A$ - complex linear operator\\
  $M$ - magnitude linear operator\\
  $P$ - phase linear operator\\
  $\mathcal{W}$ - set of phase wraps\\
  $N$ - number of outer iterations\\
  $K$ - number of inner iterations for magnitude and phase update\\
  \textbf{Output:}\\
  $m_N$ - reconstructed magnitude images\\
  $p_N$ - reconstructed phase images

  \textbf{Function:}
  \begin{algorithmic}[H]
    \STATE $\alpha_m = \frac{1}{\lambda_\text{max}(A^* A) \lambda_\text{max}(M^* M)}$
    \FOR{$n = 0, \hdots, N - 1$}
    \STATE // Update $m$ with fixed $p_n$
    \STATE $m_{n, 0} = m_n$
    \FOR{$k = 0, \hdots, K - 1$}
    \STATE $r_{n, k} = A^* \left(y - A  (M m_{n, k} \cdot e^{\imath  P p_n }) \right)$
    \STATE $m_{n, k+1} = \prox_{\alpha_{m} g_m} \left( m_{n, k} + \alpha_{m} \text{Re} ( M^*  (e^{-\imath P p_n }  \cdot r_{n, k} ) ) \right)$
    \ENDFOR
    \STATE $m_{n+1} = m_{n, K}$
    \STATE // Update $p$ with fixed $m_{n+1}$
    \STATE $p_{n, 0} = p_n$
    \STATE $\alpha_{n} = \frac{1}{\lambda_\text{max}(A^* A) \lambda_\text{max}(P^* P) \max(|M m_{n+1}|^2)}$
    \FOR{$k = 0, \hdots, K - 1$}
    \STATE Randomly draw $w_{n, k} \in \mathcal{W}$
    \STATE $r_{n, k} = A^* \left(y - A  (M m_{n+1} \cdot e^{\imath  P p_{n, k} }) \right)$
    \STATE $p_{n, k+1} = \prox_{\alpha_{n} g_p} \left( p_{n, k} +  w_{n, k} + \alpha_{n} \text{Im}\left( P^*  (M m_{n+1} \cdot e^{-\imath P p_{n, k} }  \cdot r_{n, k} ) \right) \right) - w_{n, k}$
    \ENDFOR
    \STATE $p_{n+1} = p_{n, K}$
    \ENDFOR
  \end{algorithmic}
\end{algorithm}

\section{Phase Cycling as an Inexact Proximal Gradient Method}
\label{sec:inexact}

In this section, we show that our proposed phase-cycling is an instance of \revised{\rnum{R2.12.34}the }inexact proximal splitting method described in Sra's paper~\cite{Sra:2012wt}. Following its result, the proposed phase cycling converges to an inexact stationary point.

Concretely, the inexact proximal splitting method in Sra's paper considers the following minimization problem:
\begin{equation}
  \begin{aligned}
    \minimize{p} f(p) + g(p)
  \end{aligned}
\end{equation}
and utilizes the following update steps:
\begin{equation}
  \begin{aligned}
    p_{n+1} = \prox_{\alpha_n g} ( p_n - \alpha_n  \nabla f (p_n) + \alpha_n e_n )
  \end{aligned}
\end{equation}
where $e_n$ is the error at iteration $n$. The results in Sra's paper~\cite{Sra:2012wt} show that the algorithm converges to a solution that is close to a nearby stationary point, with distance proportional to the iteration error $e_n$. To translate this result to phase cycling, we need to express the error in the proximal operation as additive error.

In the context of phase cycling, our objective function consists of $f(p) = \frac{1}{2} \| y - A (M m \cdot e^{\imath  P p })  \|_2^2$ and  $g(p) =  \frac{1}{|\wraps|} \sum_{w \in \wraps} g_p (p + w)$. 
Let us define the regularization function error at iteration $n$ to be $\epsilon_n(p) = g_p(p + w_n) - \frac{1}{|\wraps|} \sum_{w \in \wraps} g_p (p + w)$, then the proposed phase cycling update step can be written as:
\begin{equation}
	\begin{aligned}
		p_{n+1} = \prox_{\alpha_n g + \alpha_n \epsilon_n } \left( p_n - \alpha_n  \nabla f (p_n) \right)
	\end{aligned}
	\label{eq:spgd}
\end{equation}

Now, we recall that the proximal operator is defined as $\prox_{g}(x) = \underset{z}{\text{minimize }} \frac{1}{2} \| z - x \|_2^2 + g(z)$. Then using the first-order optimality condition, we obtain that $z^* = \prox_g(x)$ if and only if $z^* = x - \nabla g(z^*)$, where $\nabla g(z^*)$ is a subgradient of $g(z^*)$. Hence, we can rewrite equation~(\ref{eq:spgd}) as,
\begin{equation}
	\begin{aligned}
		p_{n+1} &= p_n - \alpha_n  \nabla f (p_n) - \alpha_n \nabla g(p_{n+1}) - \alpha_n \nabla \epsilon_n(p_{n+1})\\
		&= \prox_{\alpha_n g} \left( p_n - \alpha_n  \nabla f (p_n) - \alpha_n  \nabla \epsilon_n(p_{n+1}) \right)
	\end{aligned}
\end{equation}
Hence, the proposed phase cycling can be viewed as an inexact proximal gradient method with error in each iteration as $e_n = \nabla \epsilon(p_{n+1})$ and converges to a point close to a stationary point. \revised{We note that this error is often bounded in practice. For example, if we consider the $\ell 1$ wavelet penalty $g(p) = \lambda \| \Psi p \|_1$, then the gradient of $g(p)$ is bounded by $\lambda$, and hence the error $e_n$ is bounded.}

\end{appendix}


\bibliography{paper}

\newpage
\section*{List of Figures}
\begin{enumerate}
\item Illustration of forward models $y = A (M m \cdot e^{j Pp})$ for partial Fourier imaging, water fat imaging and flow imaging.\label{fig:model}
  
\item Conceptual illustration of the proposed algorithm. The overall algorithm alternates between magnitude update and phase update. Each sub-problem is solved using proximal gradient descent.\label{fig:alg_overall}
  
\item Conceptual illustration of the proposed phase cycling reconstruction technique. Phase cycling achieves robustness towards phase wraps by spreading the artifacts caused by regularizing phase wraps spatially.\label{fig:alg_phase_cycle}
  
\item An example of applying smoothing regularization on the phase image by soft-thresholding its Daubechies wavelet coefficients. While the general noise level is reduced, the phase wraps are also falsely smoothened, causing errors around it, as pointed by the red arrow. These errors accumulate over iterations and cause significant artifacts near phase wraps as shown in Figure~\ref{fig:brain_proposed}.\label{fig:phase_error}
  
\item Partial Fourier + PI + CS reconstruction results on a brain dataset for the proposed method. Similar to Supplementary Figure~\ref{fig:knee}, without phase cycling, significant artifacts can be seen in the magnitude image near phase wraps in the initial solution, as pointed by the yellow arrow.\label{fig:brain_proposed}
  
\item Partial Fourier + PI + CS reconstruction comparison results on the same brain dataset in Figure~\ref{fig:brain_proposed}. The method of Bydder shows higher error in the magnitude image compared to proposed method with phase cycling in regions pointed out by the red arrows. The method of Zhao et al. shows higher magnitude image error in general, and displays staircase artifacts in the phase image, which are common in total variation regularized images, as pointed by the yellow arrow.\label{fig:brain_compare}
  
\item Water-fat + partial Fourier + PI + CS reconstruction result on a thigh dataset. Our proposed method produces similar water and fat images on a undersampled partial Fourier dataset compared to the fully-sampled reconstruction using the method from Sharma et al.\label{fig:thigh}
  
\item Net flow rates across 4 studies for the proposed method compared with $\ell 1$-ESPIRiT, calculated across segmentations of aorta (AA) and pulmonary trunk (PT). Both $\ell 1$-ESPIRiT reconstruction and proposed reconstruction result in similar flow rates.\label{fig:dfw_quant}
  
\item Divergence-free wavelet regularized flow imaging + PI + CS reconstruction result on a 4D flow dataset, compared with $\ell 1$-ESPIRiT and $\ell 1$-ESPIRiT followed by divergence-free wavelet denoising. Visually, the proposed method reduces incoherent artifacts and noise compared to the other results, especially in regions pointed by the red arrows where there should be no fluid flow.\label{fig:dfw}
  
\item Divergence-free wavelet regularized flow imaging + PI + CS reconstruction results on a 4D flow dataset. The reconstructed magnitude image has significantly reduced blurring from partial readout, compared to the $\ell 1$-ESPIRiT result.\label{fig:dfw_pf}
  
\end{enumerate}

\newpage
\section*{Supporting Figures}
\begin{enumerate}[label=S\arabic*]
  
\item Partial Fourier + PI reconstruction results on a knee dataset. Without phase cycling, significant artifacts can be seen in the magnitude image near phase wraps in the initial solution, pointed by the yellow arrow. With phase cycling, these artifacts were reduced and the result is comparable to the robust iterative partial Fourier method with $\ell 1$ wavelet described in Bydder et al.\label{fig:knee}
  
\item Water-fat + PI + CS reconstruction result on a liver dataset with three echoes. Both the method from Sharma et al. and our proposed method produce similar water and fat images on the fully-sampled dataset and the retrospectively undersampled dataset.\label{fig:liver}
\end{enumerate}

\section*{Supporting Videos}
\begin{enumerate}[label=S\arabic*]
  
\item Reconstructed magnitude and phase images over iterations for the proposed method without phase cycling. Errors accumulate over iterations near phase wraps and cause significant artifacts in the resulting images.\label{vid:without_phase_cycling}
\item Reconstructed magnitude and phase images over iterations for the proposed method with phase cycling. With phase cycling, no significant artifacts like the ones in Supplementary Video~\ref{vid:without_phase_cycling} are seen during the iterations.\label{vid:with_phase_cycling}
\end{enumerate}

\newpage

\end{document}